%% file: main.tex
\documentclass{IEEEojcsys}

\usepackage[colorlinks,urlcolor=blue,linkcolor=blue,citecolor=blue]{hyperref}

\usepackage{color,array}

\usepackage{graphicx}

\jvol{XX}
\jnum{XX}
\paper{XX}
\pubyear{XX}
\receiveddate{XX}
\accepteddate{XX}
\publisheddate{XX}
\currentdate{XX}
\doiinfo{XX}

\usepackage[utf8]{inputenc}
\usepackage[T1]{fontenc}

\usepackage{amsthm}
\usepackage{amsmath}
\usepackage{amsfonts}
\usepackage{amssymb}
\usepackage{mathtools}
\usepackage{xurl}
\usepackage{hyperref}
\usepackage{cuted}
\usepackage{mhchem}
\usepackage{stmaryrd}
\hypersetup{colorlinks=true, linkcolor=blue, filecolor=magenta, urlcolor=cyan,}
\usepackage{mathrsfs}
\usepackage{bbold}
\usepackage{graphicx}
\usepackage[export]{adjustbox}
\usepackage{todonotes}

\usepackage{algorithm}
\usepackage{algorithmicx}
\usepackage{algpseudocode}
\algnewcommand{\LineComment}[1]{\Statex \(\triangleright\) #1}
\makeatletter
\newcommand{\algmargin}{\the\ALG@thistlm}
\makeatother
\newlength{\whilewidth}
\algdef{SE}[parWHILE]{parWhile}{EndparWhile}[1]
{\parbox[t]{\dimexpr\linewidth-\algmargin}{%
		\hangindent\whilewidth\strut\algorithmicwhile\ #1\ \algorithmicdo\strut}}{\algorithmicend\ \algorithmicwhile}%
\algnewcommand{\parState}[1]{\State%
	\parbox[t]{\dimexpr\linewidth-\algmargin}{\strut #1\strut}}
\algnewcommand{\parRequire}[1]{\Require%
	\parbox[t]{\dimexpr\linewidth-\algmargin}{\strut #1\strut}}

\input{command}

\begin{document}

\sptitle{Special Issue: Intersection of Machine Learning with Control}

\title{Differentially Private Algorithms for Statistical Verification of Cyber-Physical Systems}

% \editor{This paper was recommended by Associate Editor F. A. Author.}

\author{Yu Wang\affilmark{1}}
\author{Hussein Sibai\affilmark{2}}
\author{Mark Yen\affilmark{1}}
\author{Sayan Mitra\affilmark{3}}
\author{Geir E. Dullerud\affilmark{4}}
\affil{Department of Mechanical \& Aerospace Engineering of the University of Florida, Gainesville, FL.} 
\affil{Department of Electrical Engineering and Computer Sciences at the University of California, Berkeley, CA.} 
\affil{Department of Department of Electrical \& Computer Engineering at University of Illinois at Urbana-Champaign, Urbana, IL.}
\affil{Department of Mechanical Science and Engineering at University of Illinois at Urbana-Champaign, Urbana, IL.}

% \thanks{Yu Wang and Mark Yen are with the Department of Mechanical \& Aerospace Engineering of the University of Florida, Gainesville, FL 32611, USA. Email: {\tt\small \{yuwang1, markyen\}@ufl.edu}. 
% Hussein Sibai is with the Department of Electrical Engineering and Computer Sciences at the University of California, Berkeley, CA. Email: {\tt\small hsibai@berkeley.edu}. Sayan Mitra is with the Department of Electrical \& Computer Engineering at the University of Illinois at Urbana-Champaign, IL 61801, USA. Email: {\tt\small mitras@illinois.edu}. 
% Geir E. Dullerud is the Department of Mechanical Science and Engineering of the University of Illinois at Urbana-Champaign, IL 61801, USA. Email: {\tt\small dullerud@illinois.edu}.}}

\corresp{CORRESPONDING AUTHOR: Yu Wang (e-mail: \href{mailto:yuwang1@ufl.edu}{yuwang1@ufl.edu})}
% \authornote{This work was supported by the Canada Research Chair CRC and the National Research and Engineering Council of Canada NSERC.}

\markboth{Differentially Private Algorithms for Statistic	al Verification of Cyber-Physical Systems}{Yu Wang {\itshape et al}.}

% \author{Yu Wang, Hussein Sibai, Mark Yen, Sayan Mitra, and Geir E. Dullerud% <-this % stops a space
\input{abs}

\maketitle

\input{intro}

\input{prelim}

\input{privacy}

\input{adp}

\input{algorithm}

\input{case}

\input{conc}

\bibliography{yu}
\bibliographystyle{IEEEtran}

\end{document}

%% file: command.tex
\newtheorem{definition}{Definition}
\newtheorem{assumption}{Assumption}
\newtheorem{theorem}{Theorem}
\newtheorem{remark}{Remark}
\newtheorem{example}{Example}
\newtheorem{lemma}{Lemma}

\renewcommand{\d}{\mathrm{d}}

\newcommand{\nat}{\mathbb{N}}
\newcommand{\real}{\mathbb{R}}
\newcommand{\nnreal}{\real_{\geq 0}}

%% file: abs.tex
\begin{abstract}
Statistical model checking is a class of sequential algorithms that can verify specifications of interest on an ensemble of cyber-physical systems (e.g., whether 99\% of cars from a batch meet a requirement on their energy efficiency). These algorithms infer the probability that given specifications are satisfied by the systems with provable statistical guarantees by drawing sufficient numbers of independent and identically distributed samples. During the process of statistical model checking, the values of the samples (e.g., a user's car energy efficiency) may be inferred by intruders, causing privacy concerns in consumer-level applications (e.g., automobiles and medical devices). This paper addresses the privacy of statistical model checking algorithms from the point of view of differential privacy. These algorithms are sequential, drawing samples until a condition on their values is met. We show that revealing the number of samples drawn can violate privacy. We also show that the standard exponential mechanism that randomizes the output of an algorithm to achieve differential privacy fails to do so in the context of sequential algorithms. Instead, we relax the conservative requirement in differential privacy that the sensitivity of the output of the algorithm should be bounded to any perturbation for any data set. We propose a new notion of differential privacy which we call \emph{expected differential privacy}. Then, we propose a novel expected sensitivity analysis for the sequential algorithm and proposed a corresponding exponential mechanism that randomizes the termination time to achieve the expected differential privacy. We apply the proposed mechanism to statistical model checking algorithms to preserve the privacy of the samples they draw. The utility of the proposed algorithm is demonstrated in a case study.
\end{abstract}

\begin{IEEEkeywords}
cyber-physical systems, stochastic systems, formal verification, statistical model checking, privacy
% Enter key words or phrases in alphabetical order, separated by commas. 
% For a list of\goodbreak suggested keywords, send a blank e-mail to \href{mailto:keywords@ieee.org}{keywords@ieee.org} or visit\goodbreak \href{http://www.ieee.org/organizations/pubs/ani_prod/keywrd98.txt}{http://www.ieee.org/organizations/pubs/ani\_prod/keywrd98.txt}
\end{IEEEkeywords}

%% file: intro.tex
\section{Introduction} 
\label{sec:intro}

Cyber-physical systems appear naturally when physical processes are controlled by computer-based algorithms~\cite{rajkumar2010cyberphysical} in applications such as include automobiles~\cite{jin2014benchmarks}, smart grids~\cite{daniele2017smart}, and medical/health devices~\cite{lee2010medical}. These systems are typically designed in a compositional fashion by interconnecting numerous cyber and physical components. In operation, they are subject to variability from different sources, e.g., control algorithm updates and physical wear and tear. In these contexts, besides pursuing correct-by-construction, verification plays an important role in assuring the systems' functionality in critical applications~\cite{rajkumar2010cyberphysical}.

Statistical model checking is a commonly-used class of verification algorithms that can verify general specifications for (an ensemble of) cyber-physical systems. These specifications are formally expressed using temporal logic, which is composed of a simple set of syntactic rules. These syntactic rules augment the standard propositional logic with a few temporal operators.
%, alongside propositional logical operators ``not'', ``and'', and ``or''. 
For any temporal logic specification of interest, the statistical model checker can automatically parse it by the semantic rules of the logic and divide it into several sub-specifications, which are verifiable by basic statistical inference techniques. Although the verification results are subject to statistical errors, these are usually tolerable for most applications~\cite{agha2018survey}. Compared to other model-based verification methods (e.g.,~\cite{alur2015principles,coogan2015efficient,roohi2017hare,sibai2019using}), statistical model checking  is more scalable and can handle black-box systems. They have been successfully used to verify various specifications of real-world cyber-physical systems from automobiles~\cite{jin2014benchmarks,barbot2017statistical,wang2019statisticala} and system biology~\cite{zuliani2015statistical}.

When statistical model checking algorithms infer specifications for (an ensemble of) cyber-physical systems, the values of the samples they draw may be compromised by intruders by them observing their outputs and termination times, as is the case of other statistical inference algorithms~\cite{dwork2006differential}. Such an issue causes privacy concerns of statistical model checking in consumer-level applications (e.g., automobiles and health/medical devices) where the system samples are related to sensitive personal information.

To prevent such privacy violations, the concept of differential privacy has been proposed and adopted by various industries, e.g., Google~\cite{erlingsson2014rappor}, Apple~\cite{appledifferentialprivacyteam2017learning}, and the U.S. Census Bureau~\cite{abowd2020modernization}. Differential privacy requires the (random) observations of an algorithm not to change significantly in the probabilistic sense when small changes are made to its sample values. If such a bound on the change exists, it will subsequently provide privacy guarantees on how hard it is to infer the values of the samples from the observations.

The main goal of this work is to develop a new statistical model checking algorithm that can preserve the differential privacy of the samples used. 
%A motivating example is as follows. 
For example, consider a set of cars manufactured by a certain company. We model their energy-efficiency levels by a probabilistic model. We are interested in checking whether at least $99\%$ of them are energy-efficient. A car is energy efficient if it can drive at least $32$ miles-per-gallon (MPG) when its speed is between $70-80$ miles per hour. A statistical model checking algorithm can check this specification by randomly sampling cars and checking whether their driven MPG satisfies the specification. The algorithm samples cars until it can conclude whether the specification is satisfied or not with a given confidence level. Since the verification result and the number of cars sampled are expected to be publicly released, there is a risk that the privacy of whether a sampled car is energy efficient could be violated. The intruder can infer whether a specific car is energy-efficient by observing how the verification result and the number of cars change with/without that car being sampled.

The privacy in this context differs from most existing literature on differential privacy in two aspects. First, existing literature focuses on algorithms that use a fixed number of data entries in their inference~\cite{dwork2013algorithmic}. For these algorithms, protecting the data privacy when publicly releasing their outputs can be done by analyzing their outputs' sensitivity to changes in their input data and accordingly randomizing the latter in what is called the  {\em exponential mechanism}~\cite{mcsherry2007mechanism}. However, statistical model checking algorithms are sequential. They draw a varying number of samples depending on a given termination condition. And releasing their verification results and the number of samples they draw may endanger privacy. We will show in Section~\ref{sub:Laplace} that the sensitivity of the number of samples, i.e., the maximal difference caused by changing one sample, can be arbitrarily large. It implies that the standard exponential mechanism, which applies to algorithms with finite sensitivity to changes in the input samples for any possible values of these samples, fails to achieve standard differential privacy. Although there have been previous works on differential privacy for sequential algorithms~\cite{ghassemi2016differentially,jain2011differentially,tsitsiklis2018private}, they usually assume that the number of samples used is not observable.

The second difference from previous work is that the data used by statistical model checking algorithms are independently and identically distributed (i.i.d.) samples from a probabilistic model instead of taking arbitrary values. Consequently, for the sequence of samples used for statistical model checking, their empirical distribution converges to the underline distribution of the probabilistic model. This motivates us to propose an expected version of differential privacy which we call {\em expected differential privacy}. This new definition takes into account the distribution of the data sampled by the statistical model checking algorithm instead of the standard worst-case assumption of arbitrary valued databases in standard differential privacy. The idea of utilizing the distribution of input data in differential privacy has been considered  in~\cite{yang2015bayesian}. However, that work only utilizes the distribution of part of the data in a fixed-size database. 
% Instead, we consider the distribution of sequence of samples.

The difference between the proposed expected differential privacy and the standard one is as follows. Consider a sequential algorithm $\mathscr{A}$ that takes an infinite sequences of samples $\sigma = (\sigma_i, \sigma_{-i})$ as input and generates an output $o = \mathscr{A} (\sigma)$, where the latter is publicly released. Here, $\sigma_i$ denotes an arbitrary $i^{\mathit{th}}$ sample and $\sigma_{-i}$ denotes the rest of the samples. If $\mathscr{A}$ achieves the standard differential privacy, then changing the value of $\sigma_i$ should only change the output $o$ slightly for {\em any} possible values of the samples in $\sigma_{-i}$. Instead, if $\mathscr{A}$ achieves expected differential privacy, then changing the value of $\sigma_i$ should only slightly change the average of the output $o$ over the distribution of the samples in $\sigma_{-i}$ (i.e., the average case).
%, assuming each sample in $\sigma_{-i}$ is drawn independently from the probabilistic model of interest. 

This paper constructively shows that achieving expected differential privacy for statistical model checking algorithms is feasible. Statistical inference in statistical model checking is performed by sequential probability ratio tests~\cite{wald1945sequential}. 
We propose a new exponential mechanism that randomizes the ratio tests using a novel type of sensitivity analysis which we call {\em expected sensitivity analysis}. Then, we develop a new statistical model checking algorithm for general signal temporal logic specifications with expected differential privacy guarantees. We demonstrate the scalability and applicability of the modified algorithm in a case study on the Toyota powertrain system.

The rest of the paper is organized as follows. Section~\ref{sec:prelim} provides some preliminaries on statistical model checking. Section~\ref{sec:differential privacy} explains why achieving standard differential privacy for sequential algorithms is difficult. Section~\ref{sec:expected differential privacy} proposes the new notion of expected differential privacy. Section~\ref{sec:algorithm} develops an expectedly differentially private statistical model checking algorithm. Section~\ref{sec:case} provides a case study on the Toyota powertrain system. Finally, Section~\ref{sec:conc} concludes this work.

\paragraph*{Notations} We denote the set of natural, real numbers, and non-negative real numbers by $\nat$, $\real$ and $\nnreal$, respectively. For $n \in \nat$, let $[n] = \{1, \ldots, n\}$. The binomial distribution has a probability mass function in the following form
$$f(K; N, p) = \frac{N!}{K! (N-K)!} p^K (1-p)^{N-K},$$ 
and we denote it by $\mathrm{Binom} (N, p)$, where $K, N \in \nat$ and $p$  is a real value in the interval $[0,1]$. The exponential distribution has a probability density function of the form 
$$
f(x) = \begin{cases}
\varepsilon e^{- \varepsilon x}, & x \geq 0 \\
0, & x < 0. \label{eq:L}
\end{cases}
$$
and we denote it by $\mathrm{Exp} (\varepsilon)$, where $\varepsilon \in \nnreal$.

%% file: prelim.tex
\section{Preliminaries on Statistical Model Checking} \label{sec:prelim}

Real-world cyber-physical systems are typically subject to uncertainty in their parameters from different sources, e.g., thermal fluctuations in sensors/actuators. The probabilistic uncertainty is expressible by random variables drawn from probabilistic distributions, which may be unknown. For example, the uncertainty in the sensor readings can be expressed by the addition of a random noise. Such uncertainties can be catpured by a probabilistic model. A key problem for the probabilistic model is whether an specification holds on a random system path with probability above some given threshold. Such a problem can be solved with provable probabilistic guarantees by statistical model checking. Below, we summarize previous work on statistical model checking algorithms based on sequentially probability ratio tests~\cite{legay2010statistical,agha2018survey}.

Consider a random signal (e.g., paths/trajectories) $\sigma$ from a probabilistic model $\mathcal{S}$. 
The model $\mathcal{S}$ can be either known or unknown and of general form (e.g., discrete, continuous, or hybrid).
% \todo{should we define probabilistic systems more formally?}
The goal of statistical model checking is to check the satisfaction probability of a specification $\varphi$. To start with, we check whether the probability of $\sigma$ satisfying $\varphi$ is greater than some given probability threshold $p \in [0,1]$, i.e.,
\begin{equation} \label{eq:problem 1}
	\mathbb{P}_{\sigma \sim \mathcal{S}} ( \sigma \models \varphi ) = p_\varphi > p,
\end{equation}
where $\models$ means ``to satisfy'' and $p_\varphi$ is the satisfaction probability of $\varphi$ on $\mathcal{S}$.

Formally, the specifications are typically expressed by signal temporal logic (STL)~\cite{maler2004monitoring}, which contains a simple set of syntactic rules to describe the change of real-valued system variables in continuous time. The syntax of STL specifications is defined recursively by:
\[
\varphi :=
f (\sigma) > 0
\mid \neg \varphi
\mid \varphi \land \varphi
\mid \varphi \ \mathcal{U}_{[t_1,t_2]} \ \varphi,
\]
where $\sigma: \nnreal \to \real^n$ is a signal (i.e., a vector of real-valued system variables that changes over time), $f: \real^n \to \real$ is a real-valued function of the signal, and $[t_1, t_2]$ is a time interval with $t_2 > t_1 \geq 0$. Strictly speaking, $t_1$ and $t_2$ should be rational-valued or infinity. By recursively applying these syntactic rules, we can write arbitrarily complex STL specifications to express general specifications of interest.

Given any STL specification $\varphi$, we can define whether a signal $\sigma$ satisfies it or not, written as $\sigma \models \varphi$ or $\sigma \not\models \varphi$, using the following semantic rules: 
\begin{align*}
& \sigma \models f (\sigma) > 0  & \ \Leftrightarrow \ &   f(\sigma(0)) > 0 \\
& \sigma \models \neg \varphi  & \ \Leftrightarrow \ &  \sigma \not\models \varphi \\
& \sigma \models \varphi \land \psi  & \ \Leftrightarrow \ &  \sigma \models \varphi \land \sigma \models \psi \\
& \sigma \models \varphi \ \mathcal{U}_{[t_1,t_2]} \ \psi & \ \Leftrightarrow \ &  \exists t \in [t_1, t_2] \text{ such that } \sigma^{(t)} \models \psi \\
& & & \land \forall t' < t, \sigma^{(t')} \models \varphi
\end{align*}
where $\sigma^{(t)}$ denotes the $t$-shift of $\sigma$, defined by $\sigma^{(t)}(t') = \sigma(t + t')$ for any $t' \in \nnreal$. 
The first rule means that $\sigma$ satisfies $f (\sigma) > 0$ if and only if the initial value of $\sigma$ at time $t=0$ satisfies $f (\sigma(0)) > 0$. The fourth rule defined the ``until'' temporal operator; it means that $\sigma$ satisfies $\varphi \ \mathcal{U}_{[t_1,t_2]} \ \psi$ if and only if $\sigma$ satisfies $\varphi$ at all time instants before $t$ until it satisfies $\psi$ exactly at time $t$ for some $t \in [t_1,t_2]$.

\begin{example} \label{ex:1}
Suppose $\sigma = (\sigma_1, \sigma_2)$ where $\sigma_1$ and $\sigma_2$ are the velocity and battery level of an electric vehicle. Then the specification
$$(\sigma_1 > 5) \ \mathcal{U}_{[0, 10]} \ (\sigma_2 < 0.2)$$
means the car velocity should be greater than $5$ before the battery level drops below $0.2$ within $10$ time units. 
\end{example}

Using the STL syntax and semantics, we can define other temporal operators such as ``finally'' (or ``eventually'') and ``always'', written as $\Diamond$ and $\square$. Specifically, 
$\Diamond_{[t_1,t_2]} \varphi = \mathtt{True} \ \mathcal{U}_{[t_1,t_2]} \ \varphi$ 
means the property $\varphi$ finally holds; and 
$\square_{[t_1,t_2]} \varphi = \neg (\Diamond_{[t_1,t_2]} \neg\varphi)$ 
means the property $\varphi$ always holds. In addition, although only the inequality symbol ``$>$'' is included in the STL syntax, we can express other inequalities symbols via the semantics. For example, $f (\sigma) \geq 0$ can be expressed by $ \neg \big( {-} f (\sigma) > 0 \big)$. Below, we will also use these operators or symbols in STL specifications.

\subsection*{Sequential probability ratio test}

Statistical model checking handles the statement~\eqref{eq:problem 1} by formulating it as a hypothesis testing problem
\begin{align} \label{eq:problem 1 ht}
\begin{split}
& H_\textrm{null}: p_\varphi > p,
\\ & H_\textrm{alt}: p_\varphi \leq p, 
\end{split}
\end{align}
where $H_\textrm{null}$ and $H_\textrm{alt}$ are the null and alternative hypotheses. Then, it infers whether $H_\textrm{null}$ or $H_\textrm{alt}$ holds by drawing sample signals $\sigma_1, \sigma_2, \ldots$ from the probabilistic model $\mathcal{S}$. As with previous work~\cite{legay2010statistical,agha2018survey}, we focus on samples that are drawn independently. 

\begin{remark} \label{rem:independent}
It has been shown that correlated samples can improve the efficiency of statistical model checking~\cite{wang2018statisticala,wang2019statistical}. However,  using correlated samples requires at least partial knowledge of the system dynamics to implement. On the other hand, independent sampling can apply to general black-box systems.
\end{remark}

The statistical model checking algorithm examines the correctness of $\varphi$ on each sample signals $\sigma_i$. This process can be performed automatically by existing model checking algorithms from~\cite{maler2004monitoring,roohi2018revisiting}. With a slight abuse of notation, we represent ``True'' and ``False'' by $1$ and $0$ and define the truth values by
\begin{equation} \label{eq:varphi}
\varphi(\sigma_i) = \begin{cases}
1, & \sigma_i \models \phi, \\
0, &\text{ otherwise.}
\end{cases}
\end{equation}
Each $\varphi(\sigma_i)$ is a binary random variable that is equal to $1$ with probability $p_\varphi$. Since the sample signals are independent, the sum 
\begin{equation} \label{eq:K}
	K = \sum_{i \in [N]} \varphi (\sigma_{i})
\end{equation}
is a random variable following the binomial distribution $\mathrm{Binom} (N, p_\varphi)$. Thus, the satisfaction probability $p_\varphi$ can be approximated by the sample average $K/N$.

Our goal is to find a statistical assertion 
$\mathscr{A} \big( \sigma_1, \ldots, \sigma_N \big) \allowbreak \to \{H_\textrm{null}, H_\textrm{alt}\}$
that claims either $H_\textrm{null}$ or $H_\textrm{alt}$ holds based on the observed samples. Moreover, since $K$ is the sufficient statistics, the statistical assertion can be written as 
$$\mathscr{A} \big(K, N) \to \{H_\textrm{null}, H_\textrm{alt}\}.$$ 
Due to the randomness of $\sigma_1, \ldots, \sigma_N$, the value of the statistical assertion $\mathscr{A}$ does not always agree with the truth value of $\mathbb{P}_{\sigma \sim \mathcal{S}} (\sigma \models \varphi) < p$. To capture these probabilistic errors, we define the false positive/false negative (FP/FN) ratios as
\begin{align} 
& \alpha_\textrm{FP} = \mathbb{P}_{\sigma_1, \ldots, \sigma_N \sim \mathcal{S}} \big( \mathscr{A} = H_\textrm{alt} \mid \mathbb{P}_{\sigma \sim \mathcal{S}} (\sigma \models \varphi) > p \big), \label{eq:fp_def}
\\ & \alpha_\textrm{FN} = \mathbb{P}_{\sigma_1, \ldots, \sigma_N \sim \mathcal{S}} \big( \mathscr{A} = H_\textrm{null} \mid \mathbb{P}_{\sigma \sim \mathcal{S}} (\sigma \models \varphi) \leq p \big). \label{eq:fn_def}
\end{align}
The FP ratio is the error probability of mistakenly rejecting the null hypotheses $H_\textrm{alt}$ while~\eqref{eq:problem 1} holds. The FN ratio is the error probability of mistakenly accepting the null hypotheses $H_\textrm{null}$ while~\eqref{eq:problem 1} does not hold.

Intuitively, the assertion $\mathscr{A}$ becomes more accurate in terms of decreasing $\alpha_\textrm{FP}$ and $\alpha_\textrm{FN}$ when the number of samples $N$ increases. For any $N$, quantitative bounds of $\alpha_\textrm{FP}$ and $\alpha_\textrm{FN}$ can be derived by using either the confidence interval method~\cite{zarei2020statistical} or the sequential probability ratio test method~\cite{sen2004statistical}. The former only assumes that $p_\varphi \neq p$. The latter requires the following stronger assumption but is more efficient. This work focuses on the latter method since the indifference parameter assumption holds in most applications (e.g., in~\cite{roohi2017statistical}).

\begin{assumption} \label{as:indifference}
There exists a known indifference parameter $\delta > 0$ such that $|p_\varphi - p| > \delta$ in~\eqref{eq:problem 1}.
\end{assumption}

With Assumption~\ref{as:indifference}, it suffices to consider the two extreme cases in hypothesis~\eqref{eq:problem 1 ht} according to~\cite{sen2004statistical}, i.e.,
\begin{align} \label{eq:problem 1 sht}
& H_\textrm{null}: p_\varphi = p + \delta, \notag
\\ & H_\textrm{alt}: p_\varphi = p - \delta.
\end{align}
To distinguish between $H_\textrm{null}$ or $H_\textrm{alt}$, we consider the likelihood ratio
\begin{equation} \label{eq:probability_ratio}
\lambda(K,N) = \frac{ (p+\delta)^{K} (1-p-\delta)^{N-K} }{ (p-\delta)^{K} (1-p+\delta)^{N-K} }.
\end{equation}
Our goal is to ensure that the FP/FN ratios $\alpha_\textrm{FP}, \alpha_\textrm{FN} \leq \alpha$ for a given threshold $\alpha > 0$, which is called the desired significance level.\footnote{We choose the same threshold for $\alpha_\textrm{FP}$ and $\alpha_\textrm{FN}$ for simplicity. The method also applies to the case where two different thresholds for $\alpha_\textrm{FP}$ and $\alpha_\textrm{FN}$ are required (see~\cite{sen2004statistical}).}
By sequential probability ratio tests~\cite{casella2002statistical}, it suffices to keep on drawing samples and stop to make a statistical assertion when either of the conditions on the right-hand-side of the following equation is satisfied:
\begin{equation} \label{eq:sprt_test}
\mathscr{A} \big(K, N) = \begin{cases}
	H_\textrm{null}, &\text{ if } \lambda(K,N) \geq \frac{1 - \alpha}{\alpha}, \\
	H_\textrm{alt}, &\text{ if } \lambda(K,N) \leq \frac{\alpha}{1 - \alpha}.
\end{cases}
\end{equation}
As $N \to \infty$, by the binomial distribution, we have $\lambda(K,N) \to 0$ if $H_\textrm{null}$ holds or $\lambda(K,N) \to \infty$ if $H_\textrm{alt}$ holds, so the above procedure stops with probability $1$. This procedure can be implemented incrementally by Algorithm~\ref{alg:1}.

\begin{algorithm}[!t]
	\caption{SMC of $\mathbb{P}_{\sigma \sim \mathcal{S}} (\sigma \models \varphi) < p$.\label{alg:1}}
	
	\begin{algorithmic}[1]
		\Require Probabilistic model $\mathcal{S}$, desired significance level $\alpha$, and indifference parameter $\delta$.
		
		\State $N \gets 0$, $K \gets 0$, $\lambda \gets 1$.
		
		\While{True}
		
		\State Draw a new sample signal $\sigma$ from $\mathcal{S}$.
		
		\State $K \gets K + \varphi(\sigma)$, $N \gets N + 1$.
		
		\State $\lambda \gets \lambda \frac{ (p+\delta)^{\varphi(\sigma)} (1-p-\delta)^{1-\varphi(\sigma)} }{ (p-\delta)^{\varphi(\sigma)} (1-p+\delta)^{1-\varphi(\sigma)} }$ .
		
		\If {$\lambda \geq \frac{\alpha}{1 - \alpha}$} Return $H_\textrm{null}$
		
		\ElsIf {$\lambda \leq \frac{1 - \alpha}{\alpha}$} Return $H_\textrm{alt}$

		\Else \ Continue

		\EndIf 

		\EndWhile
		
	\end{algorithmic}
\end{algorithm}

\begin{remark} \label{rem:2}
By Assumption~\ref{as:indifference}, we do not distinguish between checking $p_\varphi > p$ or checking $p_\varphi \geq p$ in~\eqref{eq:problem 1}. In addition, by the semantics of STL in Section~\ref{sec:prelim}, checking $p_\varphi < p$ or $p_\varphi \leq p$ is equivalent to checking $p_{\neg \varphi} \geq 1 - p$ or $p_{\neg \varphi} > 1 - p$. Thus, the method for checking $p_\varphi > p$ applies to the other three cases.
\end{remark}

%% file: privacy.tex
\section{Differential Privacy in Statistical Model Checking} \label{sec:differential privacy}

Statistical model checking of the probabilistic model $\mathcal{S}$ of an ensemble of cyber-physical systems, e.g., autonomous cars, service robots, and wearable devices, requires analyzing their signals and raises privacy concerns for consumer-level applications. 
It has been shown that even when data is protected using traditional methods such as encryption and occlusion before being processed by an algorithm, an intruder may still be able to infer them by observing the algorithm's output on differing but highly similar data \cite{narayanan2008robust, mandl2021hipaa}. In this section, we recall the widely used notion of differential privacy and demonstrate that it falls short from being applicable for the data sampled by statistical model checking algorithms with their sequential decision-making behavior.

\subsection{Differential privacy for sequential algorithms} 
\label{sub:differential privacy}

We denote the statistical model checking Algorithm~\ref{alg:1} by $\mathscr{A}$. The algorithm $\mathscr{A}$ is sequential: it samples signals $\sigma_1, \sigma_2, \ldots$ from the probabilistic model $\mathcal{S}$ until its termination condition is satisfied. Namely, it stops after a sampled data-dependent number of iterations $\tau_\mathscr{A} \in \nat$ and results in an output $o_\mathscr{A} $. With a slight abuse of notation, for an input sequence of sampled signals $\sigma_{1:\infty} = (\sigma_1, \sigma_2, \ldots)$, we write the termination step and output as functions of $\sigma_{1:\infty}$ by $\tau_\mathscr{A}  (\sigma_{1:\infty})$ and $o_\mathscr{A}  (\sigma_{1:\infty})$, respectively.

Differential privacy aims to prevent malicious inference on $\sigma_{1:\infty}$. Roughly, Algorithm $\mathscr{A}$ is differentially private if an attacker cannot infer the value of $\sigma_i$ even if they can observe $(\tau_\mathscr{A} (\sigma_{1:\infty}), o_\mathscr{A}  (\sigma_{1:\infty}))$ and the values of the rest of the entries $\sigma_{-i}$ of $\sigma_{1:\infty}$, for any sequence $\sigma_{1:\infty}$ and entry $\sigma_i$ in $\sigma_{1:\infty}$.~\cite{dwork2008differential}. Clearly, for a deterministic Algorithm $\mathscr{A}$, the value of $\sigma_i$ can be inferred when the inverse of $\tau_\mathscr{A} (\cdot)$ and $o_\mathscr{A} (\cdot)$ is unique after knowing the value of $\sigma_{-i}$.

A common approach to achieve differential privacy is to randomize Algorithm $\mathscr{A}$, such that even for the same input sequence $\sigma_{1:\infty}$, the algorithm is randomly executed in slightly different ways and yields different observations. For a randomized algorithm $\mathscr{B}$ of $\mathscr{A}$, although the randomization prevents the intruder to take the inverse of $\tau_{\mathscr{B}} (\cdot)$ and $o_{\mathscr{B}} (\cdot)$, the intruder can still infer the value of $\sigma_i$ by observing the difference in $\tau_{\mathscr{B}}  (\sigma_{1:\infty})$ and $o_{\mathscr{B}}  (\sigma_{1:\infty})$ when the value of $\sigma_{-i}$ is fixed and only the value of $\sigma_i$ changes. To measure the distance between two sequences, we recall the definition of Hamming distance.

\begin{definition}
The Hamming distance between two input sequences $\sigma_{1:\infty}$ and $\sigma_{1:\infty}'$ is
$D_H (\sigma_{1:\infty},  \sigma_{1:\infty}') = n$ if and only if they are different in only $n \in \mathbb{N}$ entries. Namely, there exists indexes $1 \leq i_1 < \ldots < i_n \in \mathbb{N}$ such that
\begin{equation*}
\begin{cases}
\sigma_j = \sigma_j', & \text{ if } j \in \{i_1, \ldots, i_n\} \\
\sigma_j \neq \sigma_j', & \text{ otherwise.} 
\end{cases}
\end{equation*}
We call two input sequences $\sigma_{1:\infty}$ and $\sigma_{1:\infty}'$ adjacent if $D_H (\sigma_{1:\infty}, \sigma_{1:\infty}') = 1$.
\end{definition}

Differential privacy requires a bound on the worst case change in the probabilities of the random observations for two adjacent input sequences. If the change in distribution is smaller, then it is harder for the intruder to infer the change in the input. We apply the standard notion of differential privacy in \cite{dwork2008differential} to Algorithm~\ref{alg:1} of Section~\ref{sec:prelim} in the following definition.

\begin{definition} \label{def:differential privacy}
A randomized sequential algorithm $\mathscr{B}$ is $\varepsilon$-differentially private, if it holds that
\begin{align*} 
&\mathbb{P}_{\mathscr{B}} \Big( \big( \tau_{\mathscr{B}} (\sigma_{1:\infty}), o_{\mathscr{B}} (\sigma_{1:\infty}) \big) \in \mathcal{O} \Big)
\\ &\leq e^{\varepsilon} \mathbb{P}_{\mathscr{B}} \Big( \big( \tau_{\mathscr{B}} (\sigma_{1:\infty}'), o_\mathscr{A} (\sigma_{1:\infty}') \big) \in \mathcal{O} \Big) 
\end{align*}
for any two adjacent input sequences $\sigma_{1:\infty}$ and $\sigma_{1:\infty}'$ and any $\mathcal{O} \subseteq \mathbb{N} \times \{H_\textrm{null}, H_\textrm{alt}\}$. 
\end{definition}

The notation $\mathbb{P}_{\mathscr{B}}$ indicates that the probability is taken from the randomized algorithm $\mathscr{B}$. Similar to the standard definition of differential privacy, Definition~\ref{def:differential privacy} has the following statistical properties. First, if Algorithm $\mathscr{B}$ is differentially private, we can prove the more general statement: for any two non-adjacent input sequences $\sigma_{1:\infty}$ and $\sigma_{1:\infty}'$,
\begin{align}
& \mathbb{P}_{\mathscr{B}} \Big( \big( \tau_{\mathscr{B}} (\sigma_{1:\infty}), o_{\mathscr{B}} (\sigma_{1:\infty}) \big) \in \mathcal{O} \Big) \leq \notag 
\\ & \quad e^{\varepsilon D_H (\sigma_{1:\infty}, \sigma_{1:\infty}')} \mathbb{P}_{\mathscr{B}} \Big( \big( \tau_{\mathscr{B}} (\sigma_{1:\infty}'), o_{\mathscr{B}} (\sigma_{1:\infty}') \big) \in \mathcal{O} \Big). \label{eq:prop 0}
\end{align}
In addition, differential privacy bounds the difference in probabilities via their probability ratios, which are commonly used statistics for probabilistic inference. This bound implies that any two input sequences give exactly the same set of observations with non-zero probabilities. Formally, the condition~\eqref{eq:prop 0} implies a necessary condition for differential privacy: 
\begin{align} 
&\mathbb{P}_{\mathscr{B}} \Big( \big( \tau_{\mathscr{B}} (\sigma_{1:\infty}), o_{\mathscr{B}} (\sigma_{1:\infty}) \big) \in \mathcal{O} \Big) \neq 0 \notag
\\ & \Longleftrightarrow \mathbb{P}_{\mathscr{B}} \Big( \big( \tau_{\mathscr{B}} (\sigma_{1:\infty}'), o_{\mathscr{B}} (\sigma_{1:\infty}') \big) \in \mathcal{O} \Big) \neq 0. \label{eq:prop 1}
\end{align}
Since most statistical inference methods depend on probability ratios, the condition~\eqref{eq:prop 0} of differential privacy implies that statistical inference for the data is not easy (e.g., lower bounds on the variance of unbiased estimators~\cite{wang2017differentiala}).

\subsection{Exponential mechanism} 
\label{sub:Laplace}

By Definition~\ref{def:differential privacy}, the sequential statistical model checking Algorithm~\ref{alg:1} is not differentially private, since for any fixed input sequence $\sigma_{1:\infty}$, its termination time and output $\big( \tau_\mathscr{A} (\sigma_{1:\infty}), o_\mathscr{A} (\sigma_{1:\infty}) \big)$ are deterministic. That does not satisfy the necessary condition of differential privacy~\eqref{eq:prop 1}. The standard approach to make deterministic algorithms differentially private is to randomize their outputs~\cite{mcsherry2007mechanism}. 

For simplicity, consider only the observation of the termination time $\tau_\mathscr{A} (\sigma_{1:\infty})$. We can define the sensitivity of termination time to changes in any single entry by 
\begin{equation}
\delta_{\tau_\mathscr{A}} := \max_{D_H (\sigma_{1:\infty}, \sigma_{1:\infty}') = 1} \big| \tau_\mathscr{A} (\sigma_{1:\infty}) - \tau_\mathscr{A} (\sigma_{1:\infty}') \big|,
\end{equation}
i.e., the maximal difference in the termination time for any two adjacent input sequences of sample signals $\sigma_{1:\infty}$ and $\sigma_{1:\infty}'$.

If the sensitivity $\delta_{\tau_\mathscr{A}}$ is finite, then the following exponential algorithm achieves differential privacy. The exponential mechanism states that  randomizing algorithm $\mathscr{A}$ to $\mathscr{B}$ such that the probability of terminating at step $k$ is 
\begin{equation} \label{eq:exponential mechanism}
\mathbb{P}_{\mathscr{B}} (\tau_{\mathscr{B}} = k) = \frac{e^{- \varepsilon |\tau_\mathscr{A} - k| / \delta_{\tau_\mathscr{A}}}}{\sum_{h \in \nat} e^{- \varepsilon |\tau_\mathscr{A} - h| / \delta_{\tau_\mathscr{A}}}},
\end{equation}
where $\mathbb{P}_{\mathscr{B}}$ means that the randomness should come from the algorithm $\mathscr{B}$ itself. The exponential mechanism~\eqref{eq:exponential mechanism} is $2 \varepsilon$-differentially private~\cite{mcsherry2007mechanism}, since for any two adjacent $\sigma_{1:\infty}$ and $\sigma_{1:\infty}'$, it holds by the definition of sensitivity that for any $k \in \nat$,
\begin{equation} \label{eq:1}
\frac{e^{- \varepsilon  |\tau_\mathscr{A} (\sigma_{1:\infty}) - k| / \delta_{\tau_\mathscr{A}}}}{e^{- \varepsilon  |\tau_\mathscr{A} (\sigma_{1:\infty}') - k| / \delta_{\tau_\mathscr{A}}}} \leq e^{\varepsilon |\tau_\mathscr{A} (\sigma_{1:\infty}) - \tau_\mathscr{A} (\sigma_{1:\infty}')| / \delta_{\tau_\mathscr{A}}} \leq e^{\varepsilon},
\end{equation}
where the first inequality holds by applying the triangle inequality. Since~\eqref{eq:1} holds for any $k$, we have
\begin{equation} \label{eq:2}
\frac{\sum_{k} e^{- \varepsilon  |\tau_\mathscr{A} (\sigma_{1:\infty}) - k| / \delta_{\tau_\mathscr{A}}}}{\sum_{k} e^{- \varepsilon  |\tau_\mathscr{A} (\sigma_{1:\infty}') - k| / \delta_{\tau_\mathscr{A}}}} \leq e^\varepsilon.
\end{equation}
Thus, the probability ratio satisfies
\begin{equation} \label{eq:3}
\frac{\mathbb{P}_{\mathscr{B}} \big( \tau_{\mathscr{B}}(\sigma_{1:\infty}) = k \big)}{\mathbb{P}_{\mathscr{B}} \big( \tau_{\mathscr{B}}(\sigma_{1:\infty}') = k \big)} \leq e^{2 \varepsilon}.
\end{equation}

The above technique depends critically on the boundedness of the sensitivity $\delta_{\tau_\mathscr{A}}$, which is generally true for non-sequential algorithms~\cite{mcsherry2007mechanism}. However, this condition is violated for Algorithm~\ref{alg:1}, as shown in Example~\ref{ex:counter}.

\begin{example} \label{ex:counter}
Consider two adjacent input sequences of sample signals with
$$\varphi\big(\sigma_{1:\infty}\big) = (\overbrace{1, \ldots, 1}^{K_1}, \boxed{1},0,1,0, \ldots)$$
and
$$\varphi\big(\sigma_{1:\infty}'\big) = (\overbrace{1, \ldots, 1}^{K_1},\boxed{0},0,1,0, \ldots)$$
in Algorithm~\ref{alg:1}, where $K_1$ satisfies
\begin{equation} \label{eq:K_1}
\frac{ (p+\delta)^{K_1 + 1} }{ (p-\delta)^{K_1 + 1} } \geq \frac{1 - \alpha}{\alpha}, \qquad \frac{ (p+\delta)^{K_1} }{ (p-\delta)^{K_1} } < \frac{1 - \alpha}{\alpha}.
\end{equation}
Recall from equation~\eqref{eq:varphi} that $\varphi(\sigma_i) = 1$ if $\sigma_i \models \varphi$ and $\varphi(\sigma_i) = 0$ if $\sigma_i \not\models \varphi$. By the definition of $\lambda$ in equation~\eqref{eq:probability_ratio}, Algorithm~\ref{alg:1} stops at $\tau_\mathscr{A} (\sigma) = K_1 + 1$ for the input sequence of sample signals $\sigma_{1:\infty}$, and it stops at $\tau_\mathscr{A} (\sigma') = \infty$ for the other input sequence of sample signals $\sigma'_{1:\infty}$. This shows that the sensitivity $\delta_{\tau_\mathscr{A}} = \infty$ and the exponential mechanism cannot achieve $\varepsilon$-differential privacy for any finite $\varepsilon$. Finally, we note that the case of $\tau_\mathscr{A} (\sigma') = \infty$ happens with probability $0$, thus it does not violate the fact that the algorithm $\mathscr{A}$ stops almost surely with probability $1$.
\end{example}

%% file: adp.tex
\section{Expected differential privacy} \label{sec:expected differential privacy}

One assumption that we make in this paper that has not been considered in the standard definition of differential privacy (Definition~\ref{def:differential privacy}) is that the sequences of signals that the algorithm processes
% $\sigma_{1:\infty}$ and $\sigma_{1:\infty}'$ 
are independently drawn from an underlying probability distribution. This is the probability model $\mathcal{S}$. This is a stronger assumption than the arbitrarily-valued static databases assumed in the literature of differential privacy~\cite{dwork2008differential}.
Such a stronger assumption helps in defining a relaxed notion of privacy that is based on limiting the change of the {\em average} of the algorithm's output over the {\em distribution} of input sequences with respect to arbitrary changes of an arbitrary entry. In contrast, standard differential privacy is based on limiting the change of the algorithm's output for {\em any} input with respect to arbitrary changes of any of the input's entries. 
%most probable input sequences to an algorithm, instead of all possible inputs.
In the case of Example~\ref{ex:counter}, 
% while achieving differential privacy is challenging, 
as we will show in this section, the average termination time of Algorithm~\ref{alg:1} over the distribution of input sequences is bounded. Thus, the probability that an input leads to an infinite termination time as the one shown in Example~\ref{ex:counter} is negligible.

% Particularly for Algorithm~\ref{alg:1}, the significance level $\alpha$ of statistical model checking is usually taken to be $\ll 1$. Thus, the termination time $K_1$ of Algorithm~\ref{alg:1} in Example~\ref{eq:ex2} is $\gg 1$, so the probability of sampling the sequences in Example~\ref{ex:counter} is almost $0$. 
% by inequalities~\eqref{eq:K_1} the probability of sampling the sequences in Example~\ref{ex:counter} is almost $0$. 
% Hence, even though Example~\ref{ex:counter} could occur theoretically, it is very unlikely to happen in practice that the two sequences happen to be sampled by Algorithm~$\mathscr{A}$ and the intruder observing its outputs. This motivates us to take the probability distribution generating the sequences into account for differential privacy.  

Recall from Definition~\ref{def:differential privacy} that for a randomized sequential algorithm $\mathscr{B}$ to be differentially private, the difference in the probability of the observations $(\tau_{\mathscr{B}} (\sigma_{1:\infty}), o_{\mathscr{B}} (\sigma_{1:\infty}))$ to be bounded under the change of a single entry $\sigma_n$. That should be satisfied by $\mathscr{B}$ for any input sequence % possible values of the rest of the entries $\sigma_{-i}$ of 
$\sigma_{1:\infty}$. 
However, under the assumption in this paper, all the sample signals in $\sigma_{-n}$ are drawn independently from the probabilistic model $\mathcal{S}$. We build on this assumption and propose an expected version of the standard differential privacy (Definition~\ref{def:differential privacy}) that requires the boundedness of the sensitivity of the average output and termination time. 

\begin{definition} \label{def:expected differential privacy}
Denote $\{H_\textrm{null}, H_\textrm{alt}\}$ by $\{1, 0\}$. 
A randomized sequential algorithm $\mathscr{B}$ is $\varepsilon$-expectedly differentially private if
%for a random sequence of signals $\sigma_{1:\infty}$ from the probabilistic model $\mathcal{S}$, it holds that 
\begin{align} 
&\mathbb{P}_{\mathscr{B}} \Big( \mathbb{E}_{\sigma_{-n}} \big( \tau_{\mathscr{B}} (\sigma_{1:\infty}), o_{\mathscr{B}} (\sigma_{1:\infty}) \big) \in \mathcal{O} \Big) \notag
\\ & \leq e^{\varepsilon} \mathbb{P}_{\mathscr{B}} \Big( \mathbb{E}_{\sigma'_{-n}} \big( \tau_{\mathscr{B}} (\sigma_{1:\infty}'), o_{\mathscr{B}} (\sigma_{1:\infty}') \big) \in \mathcal{O} \Big),
\label{eq:lemma 1}
\end{align}
for any $n \in \nat$, any $\mathcal{O} \subseteq \mathbb{N} \times \{1, 0\}$, and any pair of signals $\sigma_n$ and $\sigma'_n$, where $\sigma_{-n}$ and $\sigma'_{-n}$ are random sequences with i.i.d. entries following the probabilistic model $\mathcal{S}$.
We say that $\mathscr{B}$ is $\varepsilon$-expectedly differentially private {\em in termination time} if 
\begin{align} 
\mathbb{P}_{\mathscr{B}} \Big( \mathbb{E}_{\sigma_{-n}} \tau_{\mathscr{B}} (\sigma_{1:\infty}) \in \mathcal{O} \Big) \leq e^{\varepsilon} \mathbb{P}_{\mathscr{B}} \Big( \mathbb{E}_{\sigma'_{-n}} \tau_{\mathscr{B}} (\sigma_{1:\infty}') \in \mathcal{O} \Big), \label{eq:edp termination time}
\end{align}
for any $\mathcal{O} \subseteq \mathbb{N}$, $\sigma_n$, and $\sigma_n'$. 
\end{definition}

% \begin{proof}
% We change the order of $\mathbb{E}_{\sigma_{-n}}$ and $\mathbb{P}_{\mathscr{B}}$ by Fubini's theorem~\cite{royden1988real} to obtain the following inequality
% \begin{align*} 
% & \mathbb{E}_{\sigma_{-n}} \mathbb{P}_{\mathscr{B}} \Big( \big( \tau_{\mathscr{B}} (\sigma_{1:\infty}), o_{\mathscr{B}} (\sigma_{1:\infty}) \big) \in \mathcal{O} \Big)
% \\ & \leq e^{\varepsilon} \mathbb{E}_{\sigma_{-n}} \mathbb{P}_{\mathscr{B}} \Big( \big( \tau_{\mathscr{B}} (\sigma_{1:\infty}'), o_{\mathscr{B}} (\sigma_{1:\infty}') \big) \in \mathcal{O} \Big).
% \end{align*}
% When the significance level $\alpha \to 0$, we have $\tau_{\mathscr{B}} (\sigma_{1:\infty}), \allowbreak \tau_{\mathscr{B}} (\sigma_{1:\infty}') \to \infty$. Thus, by the law of large numbers, 
% % \todo{I think we'll need more step in this}
% \begin{align*}
% & \mathbb{E}_{\sigma_{-n}} \mathbf{I} \bigg( \mathbb{P}_{\mathscr{B}} \Big( \big( \tau_{\mathscr{B}} (\sigma_{1:\infty}), o_{\mathscr{B}} (\sigma_{1:\infty}) \big) \in \mathcal{O} \Big)
% \\ &\leq e^{\varepsilon} \mathbb{P}_{\mathscr{B}} \Big( \big( \tau_{\mathscr{B}} (\sigma_{1:\infty}'), o_\mathscr{A} (\sigma_{1:\infty}') \big) \in \mathcal{O} \Big) \bigg) \to 1,
% \end{align*}
% where $\mathbf{I}$ is the indicator function, by law of large numbers. Thus, the condition of Definition~\ref{def:expected differential privacy} holds.
% \end{proof}

%% file: algorithm.tex
\section{Expected Differential Privacy for Statistical Model Checking}
\label{sec:algorithm}

In this section, we show that the termination time and output of Algorithm~\ref{alg:1} can be randomized, to achieve expected differential privacy. Our approach is based on a new sensitivity analysis and a novel exponential mechanism for randomizing the termination step.

\subsection{Analysis of termination time for statistical model checking} 
\label{sub:expected sensitivity analysis}

The termination time and the output $\big( \tau_\mathscr{A}, o_\mathscr{A} \big)$ of Algorithm~\ref{alg:1} depend on the likelihood ratio $\lambda(N,K)$ in equation~\eqref{eq:probability_ratio}. We consider the log-likelihood ratio as an easier variable for analysis:
\begin{equation} \label{eq:llr}
\Lambda_N = \ln \lambda(K,N) = K s_+ - (N - K) s_-, \text{ where}
\end{equation}
$$
s_+ := \ln \frac{ p+\delta }{ p-\delta } > 0, \quad s_- := \ln \frac{ 1-p+\delta }{ 1-p-\delta } > 0, \text{and}
$$
$$K = \sum_{i \in [N]} \varphi (\sigma_{i})$$ 
is an random variable that follows the binomial distribution $\mathrm{Binom} (N, p_\varphi)$.

\begin{lemma} \label{lem:random walk}
The log-likelihood ratio $\Lambda_N$ forms an asymmetric random walk with probabilities and step sizes being $(p_\varphi, s_+)$ and $(1 - p_\varphi, -s_-)$.
\end{lemma}

\begin{proof}
By equation~\eqref{eq:llr}, we have $\Lambda_{0} = 0$ and 
$$
\Lambda_{N+1} = \begin{cases}
\Lambda_{N} + s_+, & \text{ with probability } p_\varphi \\ 
\Lambda_{N} - s_-, & \text{ with probability } 1 - p_\varphi. \\
\end{cases}
$$
\end{proof}

By Lemma~\ref{lem:random walk}, Algorithm~\ref{alg:1} can be interpreted as a random walk that stops upon hitting the upper or lower bounds:
\begin{align} 
    B_+ := \ln \frac{1-\alpha}{\alpha} > 0, \quad B_- := - \ln \frac{1-\alpha}{\alpha} < 0. \label{eq:stopping condition}
\end{align}
From Lemma~\ref{lem:random walk}, the average step size of the random walk $\Lambda_N$ is 
% \hussein{ put $\mathcal{S}$ in the subscript of $\mathbb{E}$ ?}
\begin{align}
D := & \mathbb{E}_{\sigma_1} \Lambda_1 = p_\varphi s_+ - (1 - p_\varphi) s_- \notag
\\ & = p_\varphi \ln \frac{ p+\delta }{ p-\delta } - (1 - p_\varphi) \ln \frac{ 1-p+\delta }{ 1-p-\delta } \label{eq:Lambda_1}
\end{align} 
When $B_+ \gg 1$ (equivalently, $B_- = - B_+ \ll -1$), we can prove that if $p_\varphi > p + \delta$, then the average step size is positive and the asymmetric random walk $\Lambda_N$ will hit the upper bound $B_+$ with probability $1 - \alpha$ and the lower bound $B_-$ with probability $\alpha$. Similarly, if $p_\varphi < p - \delta$, then the average step size is negative and the asymmetric random walk $\Lambda_N$ will hit the lower bound $B_-$ with probability $1 - \alpha$ and the upper bound $B_+$ with probability $\alpha$. 

The average termination time $\tau_{\mathscr{A}} (\sigma_{1:\infty})$ satisfies the stopping time property of random processes~\cite{casella2002statistical}, i.e., 
% \hussein{citation? subscripts for expectations?}
\begin{align} 
\mathbb{E}_{\sigma_{1:\infty}} [\Lambda_{\tau_{\mathscr{A}} (\sigma_{1:\infty})}] 
& = \mathbb{E}_{\sigma_1} \Lambda_{1} \mathbb{E}_{\sigma_{1:\infty}} [ \tau_{\mathscr{A}} (\sigma_{1:\infty}) ] \notag
\\ & = D \mathbb{E}_{\sigma_{1:\infty}} [ \tau_{\mathscr{A}} (\sigma_{1:\infty}) ]. \label{eq:stopping time property}
\end{align}
% by setting $\mathbb{E}_{\sigma_{1:\infty}} \Lambda_{\tau_{\mathscr{A}} (\sigma_{1:\infty})} 
%  = B_+$ for $\mathbb{E}_{\sigma_1} \Lambda_{1} > 0$ and $ = B_-$ for $\mathbb{E}_{\sigma_1} \Lambda_{1} < 0$.
% This equation shows the expected stopping position is equal to the expected step size multiplied by the termination time. Accordingly, the expected termination time is
% \begin{align}
% \mathbb{E} \tau_{\mathscr{A}} = \frac{\mathbb{E} \Lambda_{\tau_{\mathscr{A}}}}{\mathbb{E} \Lambda_{1}} & = \frac{(1 - 2 \alpha) \ln \frac{1-\alpha}{\alpha}}{\big| p_\varphi \ln \frac{ p+\delta }{ p-\delta } - (1 - p_\varphi) \ln \frac{ 1-p+\delta }{ 1-p-\delta } \big|} \notag
% \\ & \approx \frac{\ln \frac{1-\alpha}{\alpha}}{\big| p_\varphi \ln \frac{ p+\delta }{ p-\delta } - (1 - p_\varphi) \ln \frac{ 1-p+\delta }{ 1-p-\delta } \big|} \label{eq:E tau}
% \end{align}
% for $\alpha \ll 1$. 

\subsection{Expected sensitivity analysis for statistical model checking}

Now we define the sensitivity of the average termination time $\tau_\mathscr{A} (\sigma_{1:\infty})$ to single entry perturbations in the input sequence in the following equation: 
\begin{equation} \label{eq:expected sensitivity}
\Delta_{\tau_\mathscr{A}} := \max_{n \in \mathbb{N}, \sigma_n, \sigma'_n} \big| \mathbb{E}_{\sigma_{-n}} \tau_\mathscr{A} (\sigma_{1:\infty}) - \mathbb{E}_{\sigma_{-n}'} \tau_\mathscr{A} (\sigma_{1:\infty}') \big|,
\end{equation}
where $\sigma_{-n}$ and $\sigma_{-n}'$ are two random sequences of signals obeying the distribution of the probabilistic model $\mathcal{S}$. 
We call $\Delta_{\tau_\mathscr{A}}$ the {\em expected sensitivity} of termination time. Expected sensitivity $\Delta_{\tau_\mathscr{A}}$ represents the maximal change in the average termination time over input sequences sampled from the probabilistic model $\mathcal{S}$ when arbitrarily changing the value of an arbitrary $n^{\mathit{th}}$ entry. %, assuming that the rest of the entries are randomly sampled from the probabilistic model $\mathcal{S}$. 

For 
%the statistical model checking algorithm $\mathscr{A}$
Algorithm~\ref{alg:1}, we can use equation~\eqref{eq:stopping time property} to compute the expected sensitivity as shown in the following lemmas.

\begin{lemma} \label{lem:termination time}
For $D > 0$ in equation~\eqref{eq:Lambda_1}, 
consider the random walk $\Lambda_N$ from Lemma~\ref{lem:random walk} for hitting any two absorbing bounds $A \gg \max\{s_+, s_-\}$ and $-B \ll - \max\{s_+, s_-\}$, the probability of hitting $B$ is $\approx e^{-B}$ and 
the expected termination time is given by
$$
\mathbb{E}_{\sigma_{1:\infty}} [ \tau_{\mathscr{A}} (\sigma_{1:\infty}) ] \approx \frac{ A (1 - e^{-B})  - B e^{-B} }{D}.
$$
\end{lemma}

\begin{proof}
Let $a$ and $b$ be the probabilities of hitting the bounds $A$ and $B$ respectively. We have $a+b =1$, since the random walk terminates with probability $1$. For $D > 0$, when $A \gg \max\{s_+, s_-\}$ and $-B \ll - \max\{s_+, s_-\}$, we have from~\cite{casella2002statistical} that
$$
b \approx \frac{1 - e^{-A}}{e^{B}} \approx e^{-B}.
$$
The first approximate equality is due to the discrete steps in the random walk, which is negligible when $A \gg \max\{s_+, s_-\}$ and $-B \ll - \max\{s_+, s_-\}$.
By the stopping time property~\eqref{eq:stopping time property}, we have the lemma holds.
\end{proof}

\begin{lemma} \label{lem:expected sensitivity}
The expected sensitivity of the termination time $\tau_{\mathscr{A}}$ satisfies
\begin{equation} \label{eq:Delta}
\Delta_{\tau_\mathscr{A}} = \frac{s_+ + s_-}{|D|}
= \frac{\ln \frac{ p+\delta }{ p-\delta } + \ln \frac{ 1-p+\delta }{ 1-p-\delta }}{\big| p_\varphi \ln \frac{ p+\delta }{ p-\delta } - (1 - p_\varphi) \ln \frac{ 1-p+\delta }{ 1-p-\delta } \big|},    
\end{equation}
where $s_+$ and $s_-$ are from equation~\eqref{eq:llr} and $D$ is from equation~\eqref{eq:Lambda_1}.
\end{lemma}

\begin{proof}
Without loss of generality, consider the case of $D > 0$ in equation~\eqref{eq:Lambda_1}. The case of $D < 0$ can be proved similarly.
Consider any two input sequences $\sigma_{1:\infty}$ and $\sigma_{1:\infty}'$ for Algorithm~\ref{alg:1}, where $\sigma_1$ and $\sigma_1'$ are chosen such that $\sigma_1 \models \varphi$ and $\sigma_1' \not\models \varphi$, and $\sigma_{2:\infty}$ and $\sigma_{2:\infty}'$ are drawn in an i.i.d. manner from the probabilistic model $\mathcal{S}$.
Then, for the input sequence $\sigma_{1:\infty}$, the log-likelihood ratio $\Lambda_N$ follows a random walk that starts at $s_+$ after the first step. To derive the termination time to hit either $B_+$ or $B_-$ from equation~\eqref{eq:stopping condition}, we set $A := B_+ - s_+$ and $B := B_- - s_+$ in Lemma~\ref{lem:termination time} to obtain  
% \todo{sometimes $\tau_\mathcal{A}$ is treated as a function of $\sigma$ and sometimes as a value, the notation should be unified in addition to the terminology. Also the expectations should have subscripts. Shouldn't $\Lambda_{1}$ be also a function of $\sigma$?}
\begin{align*}
& \mathbb{E}_{\sigma_{2:\infty}} [ \tau_{\mathscr{A}} (\sigma_{1:\infty}) ] 
\\ & = 1 + \frac{(B_+ - s_+) (1 - e^{B_- - s_+}) + (B_- - s_+) e^{B_- - s_+} }{D}.
\end{align*} 
Similarly, by setting $A = B_+ + s_-$ and $B = B_- + s_-$ in Lemma~\ref{lem:termination time} as the bounds for the random walk of the log-likelihood ratio $\Lambda'_N$ of $\sigma'_{2:\infty}$, we have
\begin{align*}
& \mathbb{E}_{\sigma_{2:\infty}'} [ \tau_{\mathscr{A}} (\sigma_{1:\infty}') ] 
 \\ & = 1 + \frac{(B_+ + s_-) (1 - e^{B_- + s_-}) + (B_- + s_-) e^{B_- + s_-} }{D}.
\end{align*}
Considering $s_+, -s_- \ll B_+, -B_-$, the subtraction of the two equations shown above gives
\begin{align*}
\mathbb{E}_{\sigma_{2:\infty}} \tau_{\mathscr{A}} (\sigma_{1:\infty}) - \mathbb{E}_{\sigma_{2:\infty}'} \tau_{\mathscr{A}} (\sigma_{1:\infty}') \approx \frac{s_+ + s_-}{D},
\end{align*}
considering that $e^{B_-} \ll 1$. This proves the bound of expected sensitivity~\eqref{eq:expected sensitivity} stated in the lemma for alternating the first entry. The case of other entries can be proved similarly.
\end{proof}

When the expected sensitivity $\Delta_{\tau_\mathscr{A}} < \infty$ is finite, we can apply a modified version of the exponential mechanism described in Section~\ref{sub:Laplace} to Algorithm~\ref{alg:1} to achieve expected differential privacy for the termination time. This is shown in the following lemma.

\begin{lemma} \label{lem:exponential mechanism 2}
For a sequential algorithm $\mathscr{A}$ with a deterministic termination time $\tau_\mathscr{A}$, consider another sequential algorithm $\mathscr{B}$ with the same input and output spaces as $\mathscr{A}$ but with a random termination time $\tau_{\mathscr{B}}$ that satisfies 
\begin{align} 
& \mathbb{P}_{\mathscr{B}} \big( \mathbb{E}_{\sigma_{1:\infty}} \big[ \tau_\mathscr{B} (\sigma_{1:\infty})\big] = k \big) \notag
\\ &  
\qquad = \frac{e^{- \varepsilon |k - \mathbb{E}_{\sigma_{1:\infty}} [\tau_\mathscr{A} (\sigma_{1:\infty})]| / \Delta_{\tau_\mathscr{A}}}}{\sum_{h \in \nat} e^{- \varepsilon |h - \mathbb{E}_{\sigma_{1:\infty}} [\tau_\mathscr{A} (\sigma_{1:\infty})]| / \Delta_{\tau_\mathscr{A}}}}, \label{eq:exponential mechanism 2}
\end{align}
where $\sigma_{1:\infty}$ is a random sequence of signals obeying the distribution of the probabilistic model $\mathcal{S}$. 
Then, it is  $\varepsilon$-expectedly differentially private in termination time.
\end{lemma}

\begin{proof}
Without loss of generality, assuming $D > 0$ in equation~\eqref{eq:Lambda_1}, we will prove the condition~\eqref{eq:edp termination time} of Definition~\ref{def:expected differential privacy} for $n = 1$, i.e.,
\begin{align}
\frac{\mathbb{P}_{\mathscr{B}} \big( \mathbb{E}_{\sigma_{2:\infty}} \big[ \tau_\mathscr{B} (\sigma_{1:\infty}) ] = k \big)}{\mathbb{P}_{\mathscr{B}} \big( \mathbb{E}_{\sigma_{2:\infty}'} \big[ \tau_\mathscr{B} (\sigma_{1:\infty}') ] = k )} \leq e^{2 \varepsilon}, \label{eq:to prove}
\end{align}
for any pair of signals $\sigma_1$ and $\sigma'_1$, 
where $\sigma_{2:\infty}$ and $\sigma_{2:\infty}'$ are random sequences of i.i.d. signals obeying the distribution of the probabilistic model $\mathcal{S}$. For simplicity, we consider the case of $\varphi(\sigma_1) = 1$ and $\varphi(\sigma_1') = 0$. The case of $\varphi(\sigma_1) = 0$ and $\varphi(\sigma_1') = 1$ can be handled similarly. The cases of $\varphi(\sigma_1) = \varphi(\sigma_1') = 0$ and $\varphi(\sigma_1) = \varphi(\sigma_1') = 1$ are trivial.

When $\varphi(\sigma_1) = 1$, by the stopping time property~\eqref{eq:stopping time property}, we have 
\begin{align*}
& \mathbb{P}_{\mathscr{B}} \big( \mathbb{E}_{\sigma_{2:\infty}} \big[ \tau_\mathscr{B} (\sigma_{1:\infty}) ] = k \big)
 \\ & = \mathbb{P}_{\mathscr{B}} \big( \mathbb{E}_{\sigma_{2:\infty}} \big[ \tau_\mathscr{B} (\sigma_{2:\infty}) ] = k - \frac{s_+}{D} \big)
\end{align*}
where $s_+$ is given by equation \eqref{eq:llr} with $\mathbb{E}_{\sigma_{2:\infty}} \Lambda_{\tau_{\mathscr{B}} (\sigma_{1:\infty})} - \mathbb{E}_{\sigma_{2:\infty}} \Lambda_{\tau_{\mathscr{B}} (\sigma_{2:\infty})} = s_+$.
% $$
% s_+ = \mathbb{E}_{\sigma_{1:\infty}} [\Lambda_{\tau_{\mathscr{A}} (\sigma_{1:\infty})}] - \mathbb{E}_{\sigma_{2:\infty}} [\Lambda_{\tau_{\mathscr{A}} (\sigma_{2:\infty})}] = D \mathbb{E}_{\sigma_{1:\infty}} [ \tau_{\mathscr{A}} (\sigma_{1:\infty}) ] - D \mathbb{E}_{\sigma_{2:\infty}} [ \tau_{\mathscr{A}} (\sigma_{2:\infty}) ].
% $$

By viewing $\sigma_{2:\infty}$ as a new random walk, we can apply equation \eqref{eq:exponential mechanism 2} and derive
\begin{align}
& \mathbb{P}_{\mathscr{B}} \big( \mathbb{E}_{\sigma_{2:\infty}} \big[ \tau_\mathscr{B} (\sigma_{2:\infty}) ] = k - \frac{s_+}{D} \big) \notag
\\ & = \mathbb{P}_{\mathscr{B}} \big( \mathbb{E}_{\sigma_{1:\infty}} \big[ \tau_\mathscr{B} (\sigma_{1:\infty}) ] = k - \frac{s_+}{D} + 1 \big) \notag
\\ & = \frac{e^{- \varepsilon |k - \frac{s_+}{D} + 1 - \mathbb{E}_{\sigma_{1:\infty}} [\tau_\mathscr{A} (\sigma_{1:\infty})]| / \Delta_{\tau_\mathscr{A}}}}{\sum_{h \in \nat} e^{- \varepsilon |h - \frac{s_+}{D} + 1 - \mathbb{E}_{\sigma_{1:\infty}} [\tau_\mathscr{A} (\sigma_{1:\infty})]| / \Delta_{\tau_\mathscr{A}}}} \label{eq:tau sigma}
\end{align}
Similarly, we have 
\begin{align}
& \mathbb{P}_{\mathscr{B}} \big( \mathbb{E}_{\sigma_{2:\infty}'} \big[ \tau_\mathscr{B} (\sigma_{1:\infty}') ] = k \big) \notag
\\ & = \frac{e^{- \varepsilon |k + \frac{s_-}{D} + 1 - \mathbb{E}_{\sigma_{1:\infty}'} [\tau_\mathscr{A} (\sigma_{1:\infty}')]| / \Delta_{\tau_\mathscr{A}}}}{\sum_{h \in \nat} e^{- \varepsilon |h + \frac{s_-}{D} + 1 - \mathbb{E}_{\sigma_{1:\infty}'} [\tau_\mathscr{A} (\sigma_{1:\infty}')]| / \Delta_{\tau_\mathscr{A}}}} \label{eq:tau sigma'}
\end{align}

% Other cases can be proved similarly. From equation~\eqref{eq:exponential mechanism 2}, we have
% $$ 
% \big| \mathbb{E}_{\sigma_{-n}} [\tau_\mathscr{A} (\sigma_{1:\infty})] - \mathbb{E}_{\sigma_{-n}'} [\tau_\mathscr{A} (\sigma_{1:\infty}')] \big| \leq \Delta_{\tau_\mathscr{A}},
% $$
% Since the above inequality holds for any $\sigma_n$ and $\sigma'_n$, 
% taking the expected values of $\sigma_n$ and $\sigma'_n$ on both sides and applying Jensen's inequality gives
% $$ 
% \big| \mathbb{E}_{\sigma_{1:\infty}} [\tau_\mathscr{A} (\sigma_{1:\infty})] - \mathbb{E}_{\sigma_{1:\infty}'} [\tau_\mathscr{A} (\sigma_{1:\infty}')] \big| \leq \Delta_{\tau_\mathscr{A}}.
% $$

By the triangular inequality and using the definition of $\Delta_{\tau_\mathscr{A}}$ from~\eqref{eq:Delta}, the ratio of the numerators of equations \eqref{eq:tau sigma} and \eqref{eq:tau sigma'} satisfies
\begin{equation} \label{eq:4}
e^{-\varepsilon} \leq \frac{e^{- \varepsilon |k - \frac{s_+}{D} + 1 - \mathbb{E}_{\sigma_{1:\infty}} [\tau_\mathscr{A} (\sigma_{1:\infty})]| / \Delta_{\tau_\mathscr{A}}}}{e^{- \varepsilon |k + \frac{s_-}{D} + 1 - \mathbb{E}_{\sigma_{1:\infty}'} [\tau_\mathscr{A} (\sigma_{1:\infty}')]| / \Delta_{\tau_\mathscr{A}}}} \leq e^\varepsilon.
\end{equation}
Since~\eqref{eq:4} holds for any $k$, we have
\begin{equation} \label{eq:5}
e^{-\varepsilon} \leq \frac{\sum_{h} e^{- \varepsilon |h - \frac{s_+}{D} + 1 - \mathbb{E}_{\sigma_{1:\infty}} [\tau_\mathscr{A} (\sigma_{1:\infty})]| / \Delta_{\tau_\mathscr{A}}}}{\sum_{h} e^{- \varepsilon |h + \frac{s_-}{D} + 1 - \mathbb{E}_{\sigma_{1:\infty}'} [\tau_\mathscr{A} (\sigma_{1:\infty}')]| / \Delta_{\tau_\mathscr{A}}}} \leq e^\varepsilon.
\end{equation}
Thus, by taking the ratios of~\eqref{eq:4} and~\eqref{eq:5} and applying \eqref{eq:tau sigma} and \eqref{eq:tau sigma'} results in that for any $k \in \mathbb{N}$, we can derive~\eqref{eq:to prove}.

 The condition~\eqref{eq:edp termination time} of Definition~\ref{def:expected differential privacy} for $n \neq 1$ can be proved similarly.
\end{proof}

\subsection{Causal randomization of termination time} 
\label{sub:randomizing the algorithm}

In this section, we propose a causal randomization mechanism of the termination time of Algorithm~\ref{alg:1} that achieves the probability distribution in equation~\eqref{lem:exponential mechanism 2}.

% Based on the new notion of expected sensitivity analysis, we propose a randomization mechanism for for Algorithm~\ref{alg:1}.
%, based on the exponential mechanism from Section~\ref{sub:Laplace}.
% According to the exponential mechanism from Lemma~\ref{lem:exponential mechanism 2}, suppose that for a given input sequence $\sigma_{1:\infty}$, Algorithm~\ref{alg:1} stops after $\tau_{\mathscr{A}}$ steps, then the randomization requires the original algorithm to randomly stop according to the probability distribution from~\eqref{eq:exponential mechanism 2}.  

A trivial addition of random noise to the deterministic termination time is not causal. % Consider the implementation of the randomization mechanism.
If the deterministic termination time plus the random noise is $n \leq k = \tau_{\mathscr{A}} (\sigma_{1:\infty})$, the randomized algorithm can draw additional $n - \tau_{\mathscr{A}}$ samples and ignore their values. However, if $n < k = \tau_{\mathscr{A}} (\sigma_{1:\infty})$, then the randomized algorithm $\mathscr{B}$ will not be able to causally predict that it should stop before reaching the deterministic termination time of $\mathscr{A}$. %  cannot be done causally in time. 
% there is no way the randomized algorithm can reverse the sampling
This shows the difficulty in directly randomizing the termination time of Algorithm~\ref{alg:1} to achieve expected differential privacy. 

% Our solution is to 
Instead, we propose to randomize the stopping condition of Algorithm~\ref{alg:1}. Recall from equation~\eqref{eq:stopping condition} that Algorithm~\ref{alg:1} stops upon reaching the upper or lower bounds $B_+$ and $B_-$, respectively. We modify these bounds as follows:
\begin{align}
& B_+ = \ln \frac{1-\alpha}{\alpha} + L \text{ and } \quad B_- = - \ln \frac{1-\alpha}{\alpha} - L, \text{ where }  \notag 
\\ & L \sim \textrm{Exp} \bigg( \frac{\varepsilon}{\ln \frac{ p+\delta }{ p-\delta } + \ln \frac{ 1-p+\delta }{ 1-p-\delta }} \bigg) \label{eq:randomized stopping condition}
\end{align}
The modified version of Algorithm~\ref{alg:1} is shown in Algorithm~\ref{alg:2}.
% \hussein{that achieves} expected differential privacy. 
% \todo{You're using a different definition of $L$ in Algorithm 2 than the one in (29).}

% \todo{Also, shouldn't this be a geometric distribution instead? Shouldn't L be an integer?}

\begin{lemma} \label{lem:random bounds}
The termination time of Algorithm~\ref{alg:2} satisfies the distribution shown in equation~\eqref{eq:exponential mechanism 2}.
\end{lemma}

\begin{proof}
We denote Algorithm~\ref{alg:2} by ${\mathscr{B}}$ and let $B = \ln \frac{1-\alpha}{\alpha}$. Without loss of generality, suppose that the average step size $D$ from~\eqref{eq:Lambda_1} of the random walk $\Lambda_N$ is $> 0$ . Following the randomized stopping  condition in equation~\eqref{eq:randomized stopping condition}, we have that for any $k \in \mathbb{N}$,
\begin{align*} 
& \mathbb{P}_L \big( \mathbb{E}_{\sigma_{1:\infty}} \big[ \tau_{\mathscr{B}} (\sigma_{1:\infty}) \big] = k \big)
\\ & = \mathbb{P}_L \big( \mathbb{E}_{\sigma_{1:\infty}} \big[ \Lambda_k (\sigma_{1:\infty}) \big] \geq B + L 
\\ & \qquad \text{ and } \mathbb{E}_{\sigma_{1:\infty}} \big[ \Lambda_{k - 1} (\sigma_{1:\infty}) \big] < B + L \big)
\\ & \text{[as $\mathbb{E}_{\sigma_{-n}} \big[ \Lambda_k (\sigma_{1:\infty}) \big]$ is increasing in $k$ when $D > 0$]}
\\ & = \mathbb{P}_L \big( k D \geq B + L \text{ and } (k-1) D < B + L \big)
\\ & \text{[using the stopping time property in equation~\eqref{eq:stopping time property}]}
\\ & = \int_{(k-1) D - B}^{k D - B}
\frac{\varepsilon}{\ln \frac{ p+\delta }{ p-\delta } + \ln \frac{ 1-p+\delta }{ 1-p-\delta }} e^{\frac{\varepsilon x}{\ln \frac{ p+\delta }{ p-\delta } + \ln \frac{ 1-p+\delta }{ 1-p-\delta }}} \d x
\\ & \text{[using equation~\eqref{eq:randomized stopping condition}]}
\\ & \approx \frac{D \varepsilon}{\ln \frac{ p+\delta }{ p-\delta } + \ln \frac{ 1-p+\delta }{ 1-p-\delta }} e^{\frac{\varepsilon (k D - B)}{\ln \frac{ p+\delta }{ p-\delta } + \ln \frac{ 1-p+\delta }{ 1-p-\delta }}}
\\ & \text{[using $D \ll B$]}
\\ & \propto e^{\frac{\varepsilon (k - B/D) }{\Delta_{\tau_\mathscr{A}}}} 
\\ & \text{[using~\eqref{eq:Delta}]}
\end{align*}
This leads to~\eqref{eq:exponential mechanism 2} since $\mathbb{E}_{\sigma_{1:\infty}} [\tau_\mathscr{A} (\sigma_{1:\infty})] \approx B/D$ from Lemma~\ref{lem:termination time}.
\end{proof}

Following the same process, we can apply the expected sensitivity analysis for $o_\mathscr{A}$ for Algorithm~\ref{alg:1}. 

\begin{lemma} \label{lem:obs}
The expected sensitivity $\Delta_{o_\mathscr{A}}$ of the output $o_{\mathscr{A}}$ of Algorithm~\ref{alg:1} is $\approx 0$, where $\Delta_{o_\mathscr{A}}$ is defined as 
\begin{align}
\Delta_{o_\mathscr{A}} & := \max_{n \in \mathbb{N}, \sigma_n, \sigma'_n} \big| \mathbb{E}_{\sigma_{-n}} o_\mathscr{A} (\sigma_{1:\infty}) - \mathbb{E}_{\sigma_{-n}'} o_\mathscr{A} (\sigma_{1:\infty}') \big|, \label{eq:expected sensitivity of output}
\end{align}
and $\sigma_{-n}$ and $\sigma_{-n}'$ are random sequences of i.i.d. signals obeying the distribution of the probabilistic model $\mathcal{S}$.
\end{lemma}

\begin{proof}
Consider $n = 1$. For $D > 0$, Algorithm~\ref{alg:1} returns $0$ by hitting the bound $-B$ and returns $1$ by hitting the bound $B$, as defined in Definition~\ref{def:expected differential privacy}. By Lemma~\ref{lem:termination time}, when $\sigma_1 \models \varphi$, we have $\mathbb{P}_{\sigma_{2:\infty}} [o_\mathscr{A} (\sigma_{1:\infty}) = 0] = e^{-B+s_+}$ by viewing the log-likelihood ratio $\Lambda_N$ in Algorithm~\ref{alg:1} as a random walk starting from $s_+$ after the first step, instead of starting from zero in the first step. Similarly, when $\sigma_1 \not\models \varphi$, we have $\mathbb{P}_{\sigma_{2:\infty}} [o_\mathscr{A} (\sigma_{1:\infty}) = 0] = e^{-B-s_-}$. Thus, we have
\begin{align}
\Delta_{o_\mathscr{A}} & = \max_{\sigma_1, \sigma'_1} \big| \mathbb{E}_{\sigma_{2:\infty}} [o_\mathscr{A} (\sigma_{1:\infty})] - \mathbb{E}_{\sigma_{2:\infty}'} [o_\mathscr{A} (\sigma_{1:\infty}')] \big| \notag
\\ & \leq | e^{-B+s_+} - e^{-B-s_-} | \approx 0.
\end{align}
The same analysis holds for $n \neq 1$.
\end{proof}

Thus, changing an arbitrary $n^{\mathit{th}}$ sample has almost no influence on the output of algorithm $\mathscr{A}$ in the expected sense. In sum, we present the following theorems.

\begin{theorem}
Algorithm~\ref{alg:2} is $2 \varepsilon$-expectedly differentially private.
\end{theorem}

\begin{proof}
It directly follows from Lemmas~\ref{lem:random bounds} and~\ref{lem:obs}.
\end{proof}

In addition to privacy, we provide the following results for the significance level of Algorithm~\ref{alg:2}.

\begin{theorem}
Algorithm~\ref{alg:2} has a significance level (i.e., the upper bound on the probability that this algorithm returns a wrong answer) is less than $\alpha$.
\end{theorem}

\begin{proof}
In Algorithm~\ref{alg:2}, the hitting bounds $B+L$ and $-B-L$ are always expanded versions of those of Algorithm~\ref{alg:1} since $L \geq 0$. Hence, the accuracy is improving because of drawing more samples. Specifically, following the standard analysis of sequential hypotheses testing~\cite{casella2002statistical}, for any value of $L$ in Algorithm~\ref{alg:2}, the significance level $\alpha_L$ for hitting the bounds $B+L$ and $-B-L$ satisfies 
\begin{align*}
\frac{1 - \alpha_L}{\alpha_L} = \ln B_+ \leq  \ln \frac{1-\alpha}{\alpha}
\end{align*}
since $L \geq 0$ in~\eqref{eq:randomized stopping condition}.
This implies $\alpha_L \leq \alpha$ for any value of $L$. Thus, the theorem holds.
\end{proof}

\begin{algorithm}[!t]
\caption{SMC of $\mathbb{P}_{\sigma \sim \mathcal{S}} (\sigma \models \varphi) < p$ with expected differential privacy.\label{alg:2}}

\begin{algorithmic}[1]
\Require Probabilistic model $\mathcal{S}$, desired significance level $\alpha$, and indifference parameter $\delta$, privacy level $\varepsilon$.

\State $N \gets 0$, $K \gets 0$.

\State Log-likelihood ratio $\Lambda \gets 0$.

\State $B \gets \ln \frac{1-\alpha}{\alpha}$, $L \sim \textrm{Exp} \Big( \frac{\varepsilon}{\ln \frac{ p+\delta }{ p-\delta } + \ln \frac{ 1-p+\delta }{ 1-p-\delta }} \Big)$

\While{True}

\State Draw a sample signal $\sigma$ from $\mathcal{S}$.

\State $K \gets K + \varphi(\sigma)$, $N \gets N + 1$.

\State $\Lambda \gets \Lambda + \ln \frac{ (p+\delta)^{\varphi(\sigma)} (1-p-\delta)^{1-\varphi(\sigma)} }{ (p-\delta)^{\varphi(\sigma)} (1-p+\delta)^{1-\varphi(\sigma)} }$ .

\If {$\Lambda \geq B + L$} 

\State Return $H_\textrm{null}$ 

\ElsIf {$\Lambda \leq -B - L$} 

\State Return $H_\textrm{alt}$

\Else \ Continue

\EndIf 

\EndWhile

\end{algorithmic}
\end{algorithm}

%% file: case.tex
\section{Case Study} \label{sec:case}

In this section, we apply the statistical model checking with expected differential privacy (Algorithm~\ref{alg:2}) to the Toyota Powertrain benchmark~\cite{jin2014powertrain} in MATLAB Simulink. It is composed of an air-to-fuel (A/F) ratio controller and a model of a four-cylinder spark ignition engine that includes components starting from the throttle all the way to the crankshaft. This case study focuses on the performance of the A/F ratio controller by observing the deviation percentage of the A/F ratio $\mu(t)$ from a reference A/F ratio $\mu_\text{ref}$ as follows $$e_{A/F}(t) := \frac{\mu(t) - \mu_\text{ref}}{\mu_\text{ref}}$$ 
over a simulation time horizon of $T$ seconds. The engine's revolutions per minute (RPM) speed %is subject to Gaussian noise:
follows the Gaussian distribution
\begin{equation*}
    r \sim \text{Gaussian}(r_0,\sigma^2),
\end{equation*}
where $r_0 = 1600$, and $\sigma^2=1600$. The requirement for $e_{A/F}$ is to enter a desired region $|e_{A/F}|<0.05$ within the time interval $[0.8,T]$ \cite{wang2021probabilistic}. For a given powertrain $\mathcal{S}$, we want to check if this requirement holds with probability greater than a desired threshold $p$.
% \todo{be more general and say ``some given threshold'' here?}
Formally, in the STL syntax introduced in Section~\ref{sec:prelim}, we are interested in checking the following property:
\begin{equation} \label{eq:powertrain}
    \mathbb{P}_\sigma \big( \sigma \models \Box_{[0.8, T]}(|e_{A/F}|<0.05) \big) > p,
\end{equation}
where $T=1$ and $p = 0.73$. Evaluations were performed on a desktop with Intel Core i7-10700 CPU @ 2.90 GHz and 16 GB RAM.

\textit{Results Analysis.}
We applied Algorithm~\ref{alg:2} to analyze the Toyota Powertrain with different combinations of the significance level, indifference parameter, and privacy parameters: $\alpha \in \{0.01, 0.05\}$, $\delta \in \{0.01, 0.03\}$, and $\varepsilon \in \{0.01, 0.05\}$, respectively. 

We estimated the satisfaction probability $p_\varphi \approx 0.84$ with a standard deviation of about $0.04$ using $10000$ random samples. Thus, Assumption \ref{as:indifference} for implementing Algorithm~\ref{alg:2} holds since $p_\varphi - p \approx 0.11$,
%\todo{approximate equality, since 0.84 is estimated? How do we know 0.11 is larger than the estimation error?}
which is approximately $2.75$ times the standard deviation and is greater than both values of $\delta$ considered. In addition, we know from the estimated $p_\varphi$ that the STL specification \eqref{eq:powertrain} is true,
% \hussein{how do we know? assumption? or given?}, 
so Algorithm \ref{alg:2} should return the $\text{null}$ hypothesis $(H_{\text{null}})$ with probability at least $1 - \alpha$. We ran Algorithm~\ref{alg:2} $N = 10^4$ times for each of the considered combination of parameters. Then, we calculated the algorithm's accuracy for each combination. The accuracy is defined as follows:
\begin{equation*}
    \text{Acc.} := \frac{1}{N}\sum_{i=1}^{N} \mathbf{I}( o_i = H_{\text{null}}),
\end{equation*}
where $\mathbf{I}$ is the indicator function and $o_i$ is the output of the $i^\mathit{th}$ run. We  also computed the average number of samples before termination (Sam. or $\tau_\mathscr{B}$), the average computation time in seconds (Time), and the predicted hypothesis ($H_{\text{null}}$ or $H_{\text{alt}}$). The results with 99\% confidence level for eight of these runs are shown in Table \ref{tb: powertrain}.

\begin{figure}[ht]
    \centering
    \includegraphics[width=\linewidth]{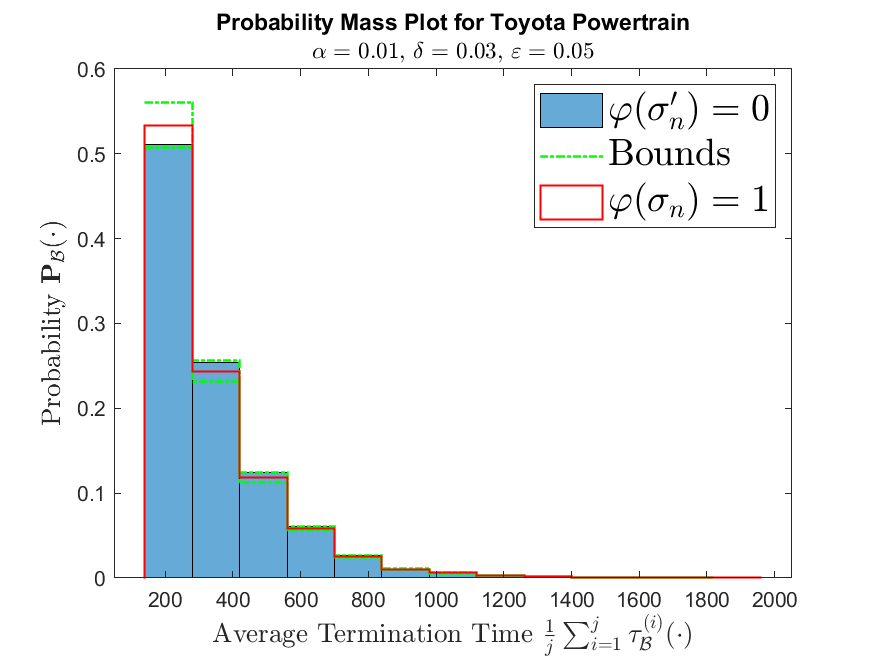}
    \caption{Histograms/empirical distributions (bin width $=140$) of the random termination time $\tau_\mathscr{A}$ for two input sequences of signals. The closeness of the two histograms indicates differential privacy.}
    \label{fig:powertrain probability mass plot 1}
\end{figure}

We then analyzed the differential privacy of Algorithm \ref{alg:2} by considering $M$ pairs of sequences of samples $\sigma_{1:\infty}$ and $\sigma_{1:\infty}'$ that differed in the $n^{\mathit{th}}$ entry.
Each entry in a sequence represents the satisfaction of $\varphi$ for the sampled execution, as defined in equation~\eqref{eq:varphi}. The average termination time over $M$ pairs of sequences is as follows:
%\todo{Use K=500 for generality like before? Used K instead of j?}

\begin{equation*}
    \text{Average Termination Time} := \frac{1}{M}\sum_{i=1}^{M} \tau_{\mathscr{B}}^{(i)}(\cdot),
\end{equation*}
where $M=500$ and $\tau_{\mathscr{B}}^{(i)}(\cdot)$ represents the termination time for the $i^\mathit{th}$ sample value of $\sigma_{1:\infty}$ and $\sigma_{1:\infty}'$. The average termination time was calculated for each of $10^4$ samples of $L$, as defined in equation~\eqref{eq:randomized stopping condition}, and the resulting distribution can be seen for one of the parameter combinations in Figure \ref{fig:powertrain probability mass plot 1}, where the red border is the distribution of the average termination time of $\sigma_{1:\infty}$ with $\varphi(\sigma_n)=1$, the blue region is the distribution of the average termination time of $\sigma_{1:\infty}'$ with $\varphi(\sigma_n')=0$, and the green dashed border is the tolerated change by factors of $e^{-\varepsilon}$ and $e^\varepsilon$ for differential privacy from Definition~\ref{def:expected differential privacy}, i.e. it is the product of the probability of average termination time of $\sigma_{1:\infty}$ being in a certain bin on the $x$-axis with $e^{-\varepsilon}$ and $e^{\varepsilon}$, respectively.

% \begin{figure}[ht]
%     \centering
%     \includegraphics[width=\linewidth]{Figures/PMF_2.png}
%     \caption{Distribution of $N$ for the second hyperparameter combination.}
%     \label{fig:powertrain probability mass plot 2}
% \end{figure}

\begin{table}[!t]
\centering
\caption{Results for Statistical Model Checking of Toyota Powertrain with expected Differential Privacy} \label{tb: powertrain}
\resizebox{0.49\textwidth}{!}{
\begin{tabular}{ c c c c r r c }
$1 - \alpha$ & $\delta$ & $\varepsilon$ & Acc. & Sam. $(\times 10^3)$ & Time (sec.) & $H_{\text{null}}$\\
\hline
$0.99$ & $0.01$ & $0.01$ & $1.00$ & $1.35 \pm 0.03$ & $181.31 \pm 4.36$ & T\\
$0.99$ & $0.01$ & $0.05$ & $1.00$ & $0.61 \pm 0.01$ & $82.04 \pm 0.65$ & T\\
$0.99$ & $0.03$ & $0.01$ & $1.00$ & $1.03 \pm 0.02$ & $146.90 \pm 9.13$ & T\\
$0.99$ & $0.03$ & $0.05$ & $1.00$ & $0.33 \pm 0.01$ & $44.86 \pm 0.58$ & T\\
$0.95$ & $0.01$ & $0.01$ & $1.00$ & $1.12 \pm 0.02$ & $163.33 \pm 8.94$ & T\\
$0.95$ & $0.01$ & $0.05$ & $1.00$ & $0.45 \pm 0.01$ & $61.81 \pm 0.58$ & T\\
$0.95$ & $0.03$ & $0.01$ & $1.00$ & $1.02 \pm 0.03$ & $137.26 \pm 4.24$ & T\\
$0.95$ & $0.03$ & $0.05$ & $1.00$ & $0.28 \pm 0.01$ & $38.11 \pm 0.58$ & T
\end{tabular}
}
\end{table}
%\todo{use $1.35 \pm 0.03 \times 10^3$ instead of $1347 \pm 32$? Also, the table is a bit too wide.}

\textit{Discussion.}
In Table \ref{tb: powertrain}, the simulation corresponding to each parameter combination yielded an accuracy of $1.00$, which agrees with the confidence level $1 - \alpha$. The high accuracy 
% compared to the confidence level 
results from the fact that we chose a small indifference parameter $\delta$, which increases termination time and thus the number of samples. This can be seen by observing that in the likelihood ratio definition in equation~\eqref{eq:probability_ratio}, increasing $\delta$ (while satisfying Assumption~\ref{as:indifference}) will reduce the termination time.
% , as has been observed in~\cite{sen2004statistical}. 
% By setting the threshold probability $p$ to be three standard deviations below $p_{\varphi}$, the algorithm had a $99.7\%$ probability of sampling signals that satisfy problem (\ref{eq:problem 1}). 
% This allowed the algorithm to consistently output the null hypothesis with accuracy greater than the confidence level $1 - \alpha$.
Relaxing the confidence level from $0.99$ to $0.95$ while holding the indifference parameter $\delta$ and privacy level $\varepsilon$ constant reduced the average number of samples needed to output the null hypothesis. 
% sThis aligns with the confidence interval method presented in \cite{zarei2020statistical}, where the algorithm continuously draws new samples to compute the significance level until it becomes less than $\alpha$.
% Increasing the indifference parameter $\delta$ while holding the confidence level and privacy level constant also reduced the average number of samples needed to arrive at a hypothesis. 
% This is expected since in the likelihood ratio (\ref{eq:probability_ratio}) where increasing $\delta$ (while satisfying Assumption~\ref{as:indifference}) will reduce the termination time.

The effects of increasing privacy level while holding the confidence level and indifference parameter constant can be seen in Table \ref{tb: powertrain} and Fig. \ref{fig:powertrain probability mass plot 1}. Increasing $\varepsilon$ decreases the privacy level and concentrates the distribution of $L$ in Algorithm~\ref{alg:2}, leading to less randomness in termination time.
% % This means that the allowable change of the distribution of termination time for adjacent input sample sequences widens.
However, this comes at the expense that it becomes easier to infer user data from sample distribution and statistical model checking output. 
From Table \ref{tb: powertrain}, the average number of samples decreased as $\varepsilon$ increased. 
In Fig. \ref{fig:powertrain probability mass plot 1}, there are a few places where the distributions fall outside the bounds. This happens because of two reasons: 1) the statistical error in approximating the distribution of the termination time with samples, and 2) the deviation of sample distributions from the expected case.  
% From Lemma \ref{lem:equivalence}, differential privacy is achieved when a sufficiently large number of samples is drawn. However, it does not define how much is considered ``sufficiently large,'' therefore increasing the number of seeds will refine the sample distribution.

%% file: conc.tex
\section{Conclusion} \label{sec:conc}

This paper studied the privacy issue of statistical model checking to enable their applications in privacy-critical applications. We used differential privacy to mathematically capture the privacy level of sample system executions that are used for statistical model checking. Since the algorithms are sequential, we showed that the termination time can violate privacy and the standard exponential mechanism fails to achieve standard differential privacy. We proposed a new exponential mechanism that can achieve a new notion of privacy which we call expected differential privacy. 
% equivalent to the standard one for large number of i.i.d. samples. 
Using the exponential mechanism, we developed expectedly-differentially private statistical model checking algorithms. The utility of the proposed algorithm was demonstrated in a case study.